\documentclass[11pt]{article}

\usepackage[final]{acl}

\usepackage{times}
\usepackage{latexsym}

\usepackage{listings}
\usepackage{xcolor}
\usepackage[most]{tcolorbox}

\usepackage[T1]{fontenc}

\usepackage[utf8]{inputenc}

\usepackage{microtype}

\usepackage{inconsolata}

\usepackage{graphicx}
\usepackage{multirow}
\usepackage{booktabs}
\usepackage{amsmath}
\title{Language Models are Lazy Poets: Creative Mode Collapse at Scale}

\author{
    Jacob A. Matthews\thanks{Equal contribution.}
    and Sil Hamilton\textsuperscript{*} \\
    Cornell University \\
    \texttt{\{jam963,srh255\}@cornell.edu}
    }

\begin{document}
\maketitle
\begin{abstract}
While aligned large language models (LLMs) can successfully complete many creative tasks, their outputs tend to be highly self-similar. Though solutions to this problem of ``mode collapse'' have been proposed, comprehensive large-scale surveys of the phenomenon in specific textual genres are still lacking.
We address this gap through an extensive treatment of LLM mode collapse in poetry.
Across 15 poetic forms, 248 subject matter tags, and 16 open-weight models, we generate over 1.1 million poems.
To our knowledge, this constitutes the largest corpus of LLM-generated poetry to date. At million-poem scale, we find that LLM poetic vocabularies and rhyming abilities are substantial, but severely underutilized. In other words, LLMs are \textit{lazy} poets: despite possessing large poetic inventories, LLMs over-rely on a small number of words, opening phrases, and rhymes. Further, we find that lexical homogenization differs across models and poetic forms, noting observed rhyming vocabulary is correlated with model size.
\end{abstract}

\section{Introduction}
Users are increasingly using large language models (LLMs) to generate novel text for purposes as diverse as coding, course content, and marketing material.
But at the same time, research suggests aligned LLMs emit repetitive content at scale despite output samples appearing unique and diverse to individual users.
The literature is increasingly concerned with this phenomenon, dubbed ``mode collapse.'' Prior work has focused on minimizing mode collapse, but the phenomenon remains underdescribed.
How significant is mode collapse?
Under what conditions do models exhibit it?
Do aligned models ever \textit{not} produce collapsed output, and if so, why?
Understanding the phenomenon is a prerequisite for proposing optimal solutions.

\begin{figure}[t]
       \centering
       \includegraphics[width=1\columnwidth]{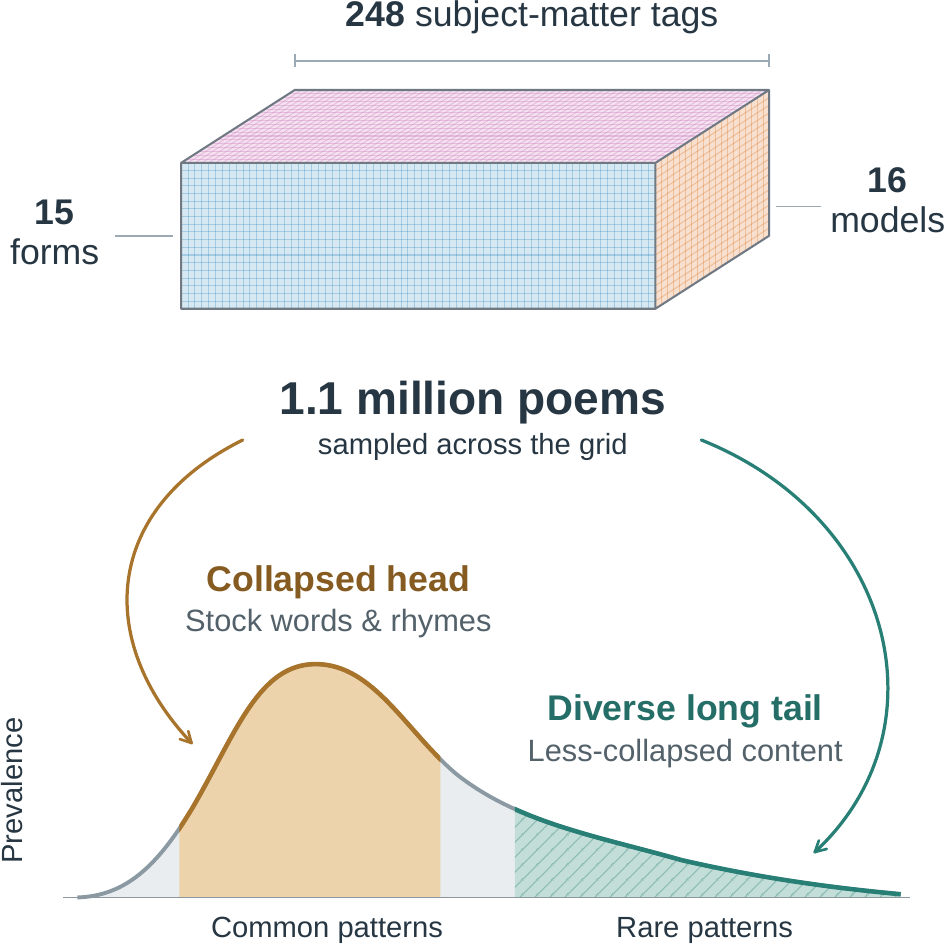}
       \caption{To investigate mode collapse at scale we generate ca. 1.1 million poems. We find that while the poems appear largely collapsed when viewed in aggregate, remarkable diversity persists in the long tail.}
       \label{fig:overview.pdf}
\end{figure}
   
In this work, we narrow the gap of understanding by generating poems at scale. 
Poems are a useful base case for studying model performance on creative tasks. Poems can be written about any subject while (often) being constrained in form.
Studying 1.1 million poems generated by 16 LLMs from five model families reveals that while models possess broad poetic capabilities, they over-rely on a small set of stock words, introductory phrases, and rhyming pairs.
In other words, LLMs are \textit{lazy} when writing poetry: their effective vocabulary is constrained relative to their observed potential.

While models do overindex on stock word types (e.g. ``light''), precisely which items a model tends to repeat depends on model family, size, and prompt content.
We find that all models possess a ``long tail'' of poetic capability and a distinctive vocabulary signature that can be obscured by self-similarity measures.
This complicates earlier work suggesting model families are converging on a shared form of mode collapse.

Our key contributions are: (i) we identify and describe model \textit{laziness} in a well-specified, open-ended creative task by generating samples at million-scale, from which (ii) we identify lexical and formal features associated with poetic ability, and finally (iii) we release all generated prompts for future researchers to employ.\footnote{Available here: \url{https://github.com/srhm-ca/LLMsAreLazyPoets}}

\section{Background}

\paragraph{Characterizing creative mode collapse.}
Early assessments of the propensity of models to emit highly typified text began shortly after the introduction of instruction tuning (and similar alignment techniques) in 2022 \citep{ouyangTrainingLanguageModels,baiConstitutionalAiHarmlessness2022,hamiltonDetectingModeCollapse}.
Analyses conducted on mode collapse found common post-training techniques improve performance in problem solving \citep{kirkUnderstandingEffectsRLHF2024} but reduce global diversity when output is observed in aggregate \citep{doshiGenerativeAIEnhances2024,zhouLargerMoreInstructable2024}.
Researchers have since proposed new post-training techniques in a bid to untangle this correlation \citep{poddarPersonalizingReinforcementLearning,zhouSequenceSequenceReward,kumarTRAININGLANGUAGEMODELS2025,liCHOICEDIVERGENCENEGLECTED2026}, but solutions have had minimal impact.
Parallel research has thus sought to characterize mode collapse across domains \citep{shumailovAIModelsCollapse2024,gerstgrasserModelCollapseInevitable2024,dohmatobTaleTailsModel2024,kovacVRecursiveTrainingLoops,wangRewardHackingEra2026}, proposing a full spectrum of inciting mechanisms including training on synthetic data and reward hacking during training.
This has led to more practical solutions including manipulating temperature during sampling \citep{troshinControlTemperatureSelective2025}, editing existing human-written text as a basis for new outputs \citep{lagzianMultiNoveltyImproveDiversity2025}, fine-tuning on high-entropy texts \citep{gambettaLearningSurpriseAdaptive2026} and, most recently, having models simply produce a range of outputs \citep{zhangVerbalizedSamplingHow2026}.
But these proposals often fail to generalize, whether because they assume model access or they require non-guaranteed input data.
Work examining the nature of mode collapse at scale is infrequent.
Casual use suggests modern LLMs do not \textit{always} emit typified text, and so we examine in what circumstances, and why, models will produce diverse output.

\paragraph{Poetry in the context of large language models.}
Poetry is a useful textual genre for testing language models because certain poetic forms (e.g., haiku, sonnets) are concretely defined and thus tractable for automated assessments of output quality via measurements including rhyming pairs and phonetic meter \citep{wockenerEndtoendStyleconditionedPoetry2021,loGPoeT2GPT2Based2022}.
Early results suggested that language models may not rhyme competently, identifying inadequacies concerning syllable-level text tokenization as one possible reason \citep{departmentofcomputersciencedartmouthcollegehanoverusaBPoMPBenchmarkPoetic2021,heuserGenerativeAestheticsFormal2025}.
Poetry has since been used as a convenient textual mode for examining mode collapse for similar reasons \citep{walshDoesChatGPTHave2024,chenEvaluatingDiversityAutomatic2024,huangLyricPoetryFace2025,heuserCulturalCollapseGenerative2025,franceschelliCreativityLargeLanguage2025,holyoakHumanArtificialCreativity2026}.
It has thus been established that models strongly collapse on particular poetic features, but existing studies have centered on heavily typified poetic forms and not less constrained forms such as free verse.
We first construct a taxonomy of poetic forms and topics to comprehensively test models and thus better characterize how mode collapse implicates rarer forms.

\section{Dataset Construction}
\begin{table}[t]
\centering
\small
\begin{tabular}{@{}p{0.60\columnwidth}cl@{}}
\toprule
Model & Year & Size (B) \\
\midrule
Qwen3 \citep{yangQwen3TechnicalReport2025} & 2025 & 4/8/14/30/32 \\
Llama 3.1 \citep{dubeyLlama3Herd2024} & 2024 & 8 \\
Llama 3.2 \citep{ai2024llama} & 2024 & 3 \\
Ministral 3 \citep{liuMinistral32026} & 2025 & 3/8/14 \\
Gemma 3 \citep{teamgemmaGemma3Technical2025} & 2025 & 4/12/27 \\
OLMo 3 \citep{allysonettingerOlmo325} & 2025 & 7 \\
OLMo 3.1 \citep{allysonettingerOlmo325} & 2025 & 32 \\
GPT-OSS \citep{openaiGptoss120bGptoss20bModel2025} & 2025 & 20 \\
\midrule
Llama 3.3 \citep{ai2024llama} & 2024 & 70 \\
GLM-4.5-Air \citep{teamGLM45AgenticReasoning2025} & 2025 & 106 \\
GPT-OSS \citep{openaiGptoss120bGptoss20bModel2025} & 2025 & 120 \\
\bottomrule
\end{tabular}
\caption{Model families and sizes considered in our experiment.
All models listed generated prompts, while those above the separator also generated poems.
In all cases we use the final instruct-tuned variant.}
\label{tab:models}
\end{table}

One method for accurately tracing how models collapse to particular output forms is to understand the space of possible outputs.
While impractical for open-ended domains like essay writing or creative fiction, certain genres of English poetry (like sonnets) are written according to formal convention.
These conventions often involve phonetic meter, rhyming patterns, line counts, and other linguistic constraints.
Topically, historical poems often also concerned a constrained set of topics including romance, nature, daily life, and so on.

\paragraph{Models.} We present the models used in our study in \autoref{tab:models}. We use only open-weight models in our study to ensure reproducibility.\footnote{We host all models on 8x RTX A6000s for 1{,}527 GPU-hours. We use the \texttt{llama.cpp} inference library at \texttt{Q8} precision and sampling parameters set to the default for each model.}

\paragraph{Taxonomy.} We construct a poetry taxonomy by drawing on \textbf{15 poetic forms and 248 subject matter topics} (``tags'') from the online Poetry Foundation corpus of English poetry cleaned and packaged by \citet{walshSonnetNotBot2024}.
We take the Cartesian product of all forms and tags to produce \textbf{3,720 unique pairs}, sparsely capturing the space of possible English poems in the public domain.

\begin{figure}[t]
    \centering
    \small
    \tcbset{
        colback=gray!6,
        colframe=gray!35,
        colbacktitle=gray!15,
        coltitle=black,
        fonttitle=\small\bfseries,
        boxrule=0.4pt,
        arc=1mm,
        left=6pt, right=6pt,
        top=5pt, bottom=5pt,
        before skip=0pt, after skip=5pt
    }
    \lstset{
        basicstyle=\small\ttfamily,
        language={},
        breaklines=true,
        breakatwhitespace=true,
        columns=fullflexible,
        keepspaces=true,
        showstringspaces=false,
        frame=none,
        literate={—}{{\textemdash}}1,
        aboveskip=0pt, belowskip=0pt
    }

    \begin{tcolorbox}[title={Prompt: Qwen3 14B}]
\begin{lstlisting}
Compose an original haiku (5-7-5 syllable
structure) that focuses on the theme of
incarceration. Ensure the poem adheres
strictly to the haiku format and centers
on the subject of incarceration. Generate
only the poem, with no additional text or
explanation.
\end{lstlisting}
    \end{tcolorbox}

    \begin{tcolorbox}[title={Response: OLMo 3.1 32B}, colback=blue!6]
\begin{lstlisting}
Cold stone walls confine
Years stretch thin in silent cells
Hope whispers softly
\end{lstlisting}
    \end{tcolorbox}

    \begin{tcolorbox}[title={Response: Llama 3.2 3B}, colback=orange!8]
\begin{lstlisting}
Cold steel walls confine
Shadows dance upon the floor
Freedom's distant dream
\end{lstlisting}
    \end{tcolorbox}

    \caption{An example prompt together with two sample responses from the dataset. Note the opening phrase is repeated almost exactly between responses, even though the generators belong to different model families.}
    \label{fig:poem_example}
\end{figure}

\paragraph{Pipeline.} To understand the dynamics of mode collapse in poem writing, we employ a data synthesis pipeline beginning with a ``metaprompt'' to instruct models to produce prompts requesting some other model to write a poem with the desired form and tag.
We provide this metaprompt in \autoref{sec:appendix-metaprompt}.
Our design minimizes the risk that models will bias their outputs toward the wording of any particular prompt, a persistent issue in mode collapse research.
This produces \textbf{70{,}680 distinct prompts} which we then use to elicit outputs from all 16 poem-generating models, yielding \textbf{1{,}130{,}888 total poems,} our final dataset.
We present sample prompts and poems in \autoref{fig:poem_example}. Prior to performing the analyses in Section \ref{sec:analysis}, we perform basic filtering, normalization, and other preprocessing to remove failed responses (such as refusals), generated titles, postscripts, and other non-poetic content. We describe this process and the results of our preprocessing in Appendix \ref{appendix:preprocessing}.

\section{Analysis}\label{sec:analysis}

\subsection{Lexical Homogenization}
\begin{figure}[t]
    \centering
    \includegraphics[width=1\columnwidth]{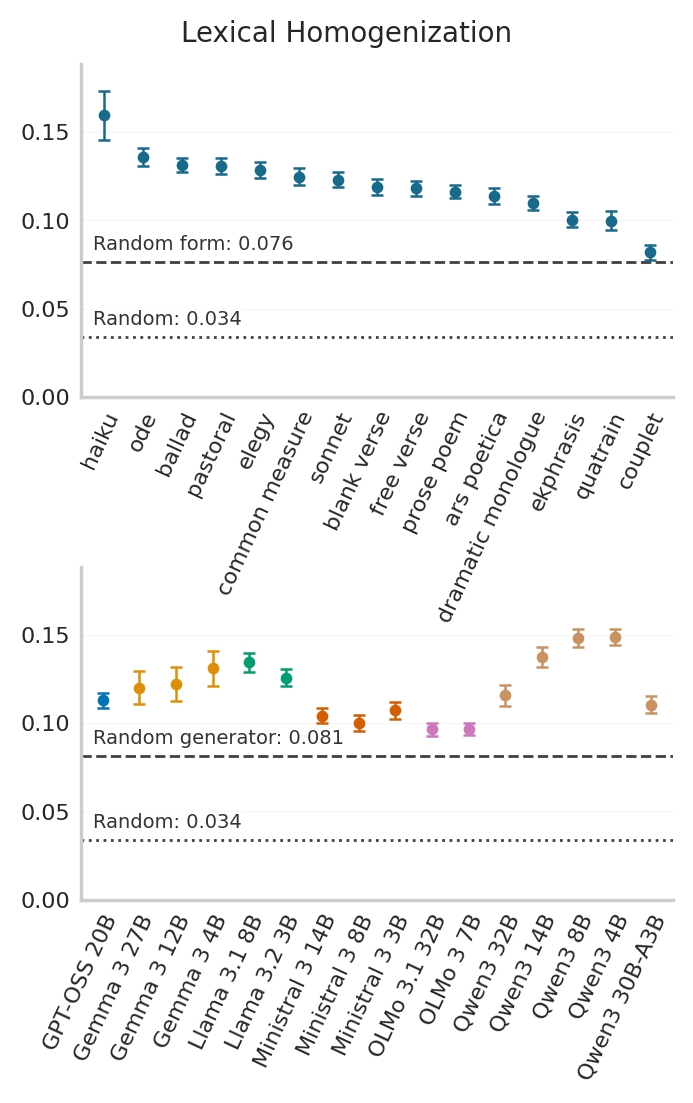}
    \caption{Mean lexical homogenization scores $\pm1$ SE, averaged by poetic form (top) and generator model (bottom), including baseline scores for randomly sampled form, generator, and 5M random pairs ("Random").}
    \label{fig:lex-homogenization}
\end{figure}

\begin{figure*}[ht]
    \centering
    \includegraphics[width=1\textwidth]{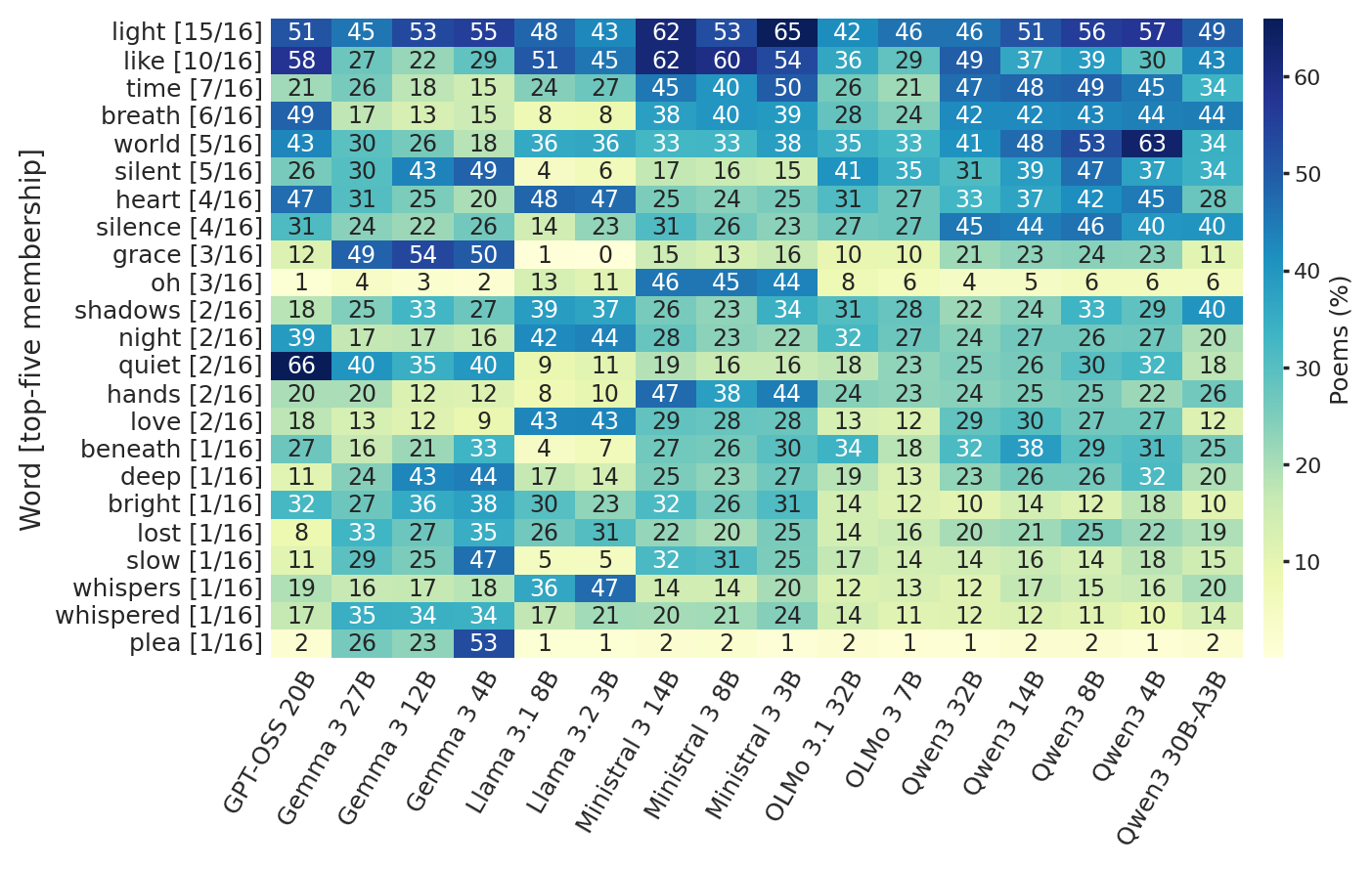}
    \caption{Union of the top-5 most commonly featured content words in generated poems. Percentages indicate share of poems by a generator that contain the listed word. The share of models with the word in their top 5 in brackets.}
    \label{fig:most-common-output-words}
\end{figure*}

 We use pairwise Jaccard similarity to measure lexical homogenization in a set of generated poems. With $W_i$ as the set of distinct content words found in poem $i$, we compute Jaccard similarity between any two poems $i$ and $j$ as 
\begin{equation}
    J(i,j)=\frac{|W_i\cap W_j|}{|W_i\cup W_j|}
\end{equation}
which expresses the share of distinct content words the poems have in common as a value between 0 and 1. Jaccard similarity measures only type presence in each poem not accounting for frequency/order. We then compute a homogeneity score $H$ for each generator $m$, form $f$, and tag $t$ as 
\begin{equation}
    H_{mft}=\frac{1}{\binom{n_{ft}}{2}}\sum_{1\le i<j\le n_{ft}} J(i, j)
\end{equation}
where $n_{ft}$ is the number of retained prompts for a given form-topic combination. If any prompt results in zero content words, unusable output, or a clear refusal from any model, that prompt is excluded for all generators; any form-tag combination with fewer than 10 valid prompts was excluded, resulting in 3,704 retained form-tag combinations. 

Using these per-cell homogeneity scores, we average across tags within each form-generator combination. We report means and standard errors across all generators for each poetic form (Figure \ref{fig:lex-homogenization}, top) and across forms for each generator (Figure \ref{fig:lex-homogenization}, bottom). For comparison, we also compute random baselines for each grouping. A random form baseline independently selects forms within each generator-tag combination, then a poem from each form; our random generator baseline independently selects generator models for the same form-tag combination. As a random baseline, we estimate the mean pairwise Jaccard between five million randomly sampled pairs of poems.  

We find that mean homogenization differs across forms: haiku rank highest ($0.159 \pm 0.014$) and couplets lowest ($0.082\pm 0.004$). All other form means lie between 0.136 and 0.10. There are also inter- and intra-model differences in within-form lexical homogenization: 
OLMo 3 7B ($0.097\pm 0.003$) and OLMo 3.1 32B ($0.096\pm 0.004$) rank lowest, while the two smallest Qwen3 models, 8B ($0.148\pm 0.005$) and 4B ($0.148\pm 0.005$) rank highest. We note that all model and form lexical homogenization scores are above both random baselines. 

\subsection{High-frequency Content Words}
In addition to lexical homogenization measures, we compute the frequency of all content words appearing in all generated poems across all models. Here, we are only interested in whether a word appears at all, not its frequency within a poem (i.e., the word's document frequency). We then identify the top-$k$ most frequent words used by each model across all poetic forms and subject matter tags. 

In Figure \ref{fig:most-common-output-words}, we report the union of the top-five most-frequently used content words for all models. Perhaps surprisingly, all models use ``light'' in more than 40\% of their generated poems. We find that the Ministral 3 series is particularly light-obsessed, with 65\% of poems by the 3B-scale model using the word. However, while ``light'' is a preoccupation shared across all models, other words are much more family-specific. Gemma 3 uses ``plea'' in between 26\% (27B) and 53\% (4B) of poems, but the same word appears in no more than 2\% of poems by any other model. Though not all models exhibit such extreme levels of lexical laziness, every model has at least one word (excluding ``light'') that appears in a third or more of its poems: for example, GPT-OSS 20B uses ``quiet'' in 66\% of outputs, ``heart'' appears in nearly half of poems by Llama 3.1 8B,  and ``world'' is present in as many as 63\% of all generations from the Qwen3 family. 

\begin{figure}[t]
    \centering
    \includegraphics[width=1\linewidth]{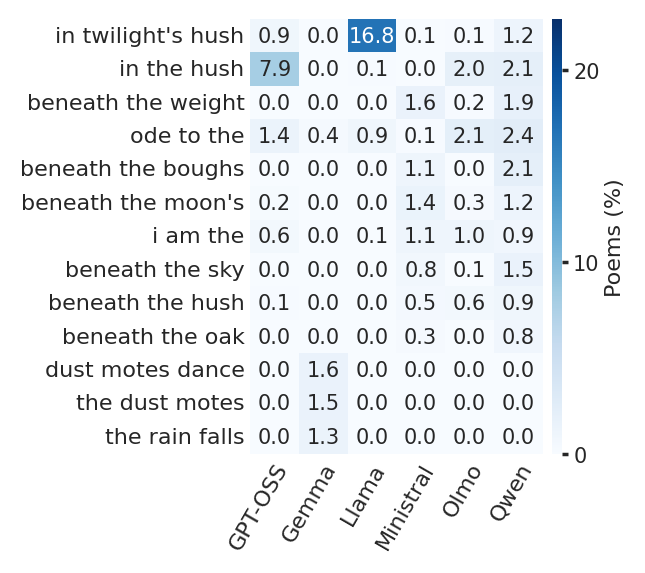}
    \caption{Percent of poems by a given model family that begin with the reported trigram. We include the union of all models' most frequent opening trigrams.}
    \label{fig:trigrams}
\end{figure}

\subsection{Recycled Openings}
In an examination of model outputs, we noted that many poems begin in very similar ways, frequently using a small set of stock phrases (see Figure \ref{fig:poem_example} for a typical example). We perform an analysis of frequent opening trigrams by extracting the most frequent three-word opening sequence for each model. Figure \ref{fig:trigrams} shows our results by model family. 

As in our lexical analysis of frequently repeated words, some models' most frequent openings are over-represented as a share of total poems. Moreover, several of these openings are themselves highly self-similar. Ministral models begin with ``in twilight's hush'' in 16.8\% of outputs, while GPT-OSS starts with ``in the hush'' in 7.9\%. Of these frequently employed incipit trigrams, six begin with the same two words: ``beneath the weight'', ``beneath the boughs'', ``beneath the moon's'', ``beneath the sky'', ``beneath the hush'', ``beneath the oak''. Oddly, Gemma 3 shows a unique preoccupation with ``dust motes'' (``the dust motes'', ``dust motes dance''). With the exception of Gemma's most frequent opening trigrams, all are found at the beginning of poems by at least two model families.\footnote{For a comparable analysis of generated poem titles, which are excluded from our main lexical analyses, see Appendix \ref{appendix:titles}.}

   \begin{table*}[t]
       \small
       \centering
       \begin{tabular*}{\textwidth}{
           @{\extracolsep{\fill}} l *{6}{c} @{}
       }
           \toprule
           & \multicolumn{6}{c}{Correlation with model size} \\
           \cmidrule(lr){2-7}
           & \multicolumn{2}{c}{Rhyming words}
           & \multicolumn{2}{c}{Rhyming pairs}
           & \multicolumn{2}{c}{Rhyming vocabulary share} \\
           \cmidrule(lr){2-3}
           \cmidrule(lr){4-5}
           \cmidrule(lr){6-7}
           Threshold
           & Corr. & $p$-value
           & Corr. & $p$-value
           & Corr. & $p$-value \\
           \midrule
           $\leq 2$     & 0.765 & 0.001 & 0.735 & 0.001 & 0.471 & 0.065 \\
           $\leq 4$     & 0.765 & 0.001 & 0.737 & 0.001 & 0.472 & 0.065 \\
           $\leq 8$     & 0.763 & 0.001 & 0.737 & 0.001 & 0.470 & 0.066 \\
           Unrestricted & 0.752 & 0.001 & 0.719 & 0.002 & 0.477 & 0.061 \\
           \bottomrule
       \end{tabular*}
       \caption{Pearson correlations between model size and rhyme
       measures at each line distance threshold. Rhyming vocabulary share
       is the number of distinct rhyming words divided by
       total observed word types for each model.}
       \label{tab:size-rhyme-correlations}
   \end{table*}

\subsection{Rhyming Vocabulary}
One of the hallmarks of many forms of English-language poetry is rhyme. While we do not measure form adherence or metrical structure here, we do examine rhyming as such. To measure rhyme, we use the CMU Pronouncing Dictionary \citep{weide1998cmudict} to identify rhyming pairs of line-terminal words. This approach is limited by the Dictionary's finite vocabulary, and it does not account for distance between line-terminal words. Because this latter limitation leads to the inclusion of ``accidental'' rhymes separated by many lines in otherwise unrhyming poems, we subset these rhyming pairs based on three line distance thresholds ($\leq 2$, $\leq 4$, $\leq 8$). For example, if a poem has rhyme scheme \textsc{ABCA}, the line-terminal words for \textsc{A} lines would be counted for thresholds 4 and 8, but not 2. Distinct pairs are counted once regardless of order (e.g. ``cat/hat'' and ``hat/cat'' count as a single pair).

With this rhyme identification approach, we count all individual words that appear in rhyming pairs, as well as all distinct pairs of rhyming words. Even under the most strict counting threshold ($\leq 2$), all models rhyme between 3,794 (Qwen3 4B) and 6,929 (OLMo 3.1 32B) distinct word types. Under the same threshold, the number of distinct pairs ranges between 9,817 (Qwen3 4B) and 20,077 (Gemma 3 27B). We report full results in Appendix Figure \ref{fig:rhyme-vocab}, and rely on this rhyming vocabulary and pair extraction data in our subsequent analyses.

\subsection{Rhyming Capability and Model Size}
To investigate the extent to which rhyming capability is related to model size, we measure Pearson correlation between all model parameter counts and three rhyming measures: number of distinct rhymed word types, number of distinct rhyming pairs, and rhyming vocabulary share. We define rhyming vocabulary share as 
\begin{equation}
   \operatorname{Share}_{m,g}
   = 100 \times \frac{|R_{m,g} \cap V_m|}{|V_m|},
\end{equation}
where $V_m$ is the set of distinct word types in poems generated by model $m$ and $R_{m,g}$ is the set of distinct words present in rhyming pairs by model $m$ under distance rule $g$. We compute correlations for all thresholds and report results in Table \ref{tab:size-rhyme-correlations}. 

Here, we find that model size is strongly ($r=0.719$ to $r=0.765$, $p\leq0.002$) correlated with the total number of distinct rhyming words and rhyming pairs. However, model size is neither strongly nor significantly ($r\leq 0.477$, $p\geq 0.061$) correlated with rhyming vocabulary expressed as a share of total observed vocabulary. When sorting models by rhyming vocabulary share (threshold $\leq2$), we find that some small models outperform larger models, even in the same family: Qwen3 14B and 8B (16.25 and 15.84\%, respectively) rank higher than the 32B variant (14.51\%), and Gemma 3's 4B variant (14.11\%) outranks the 12B version (13.39\%). This suggests that while model size is associated with increased rhyming \textit{capacity}, it has a less consistent relationship with rhyming \textit{capability}. Though larger models rhyme a greater absolute number of words and in more distinct pairs, small models are still able to rhyme a comparable share of their observed poetic vocabulary.

\begin{figure*}[ht]
    \centering
    \includegraphics[width=1\textwidth]{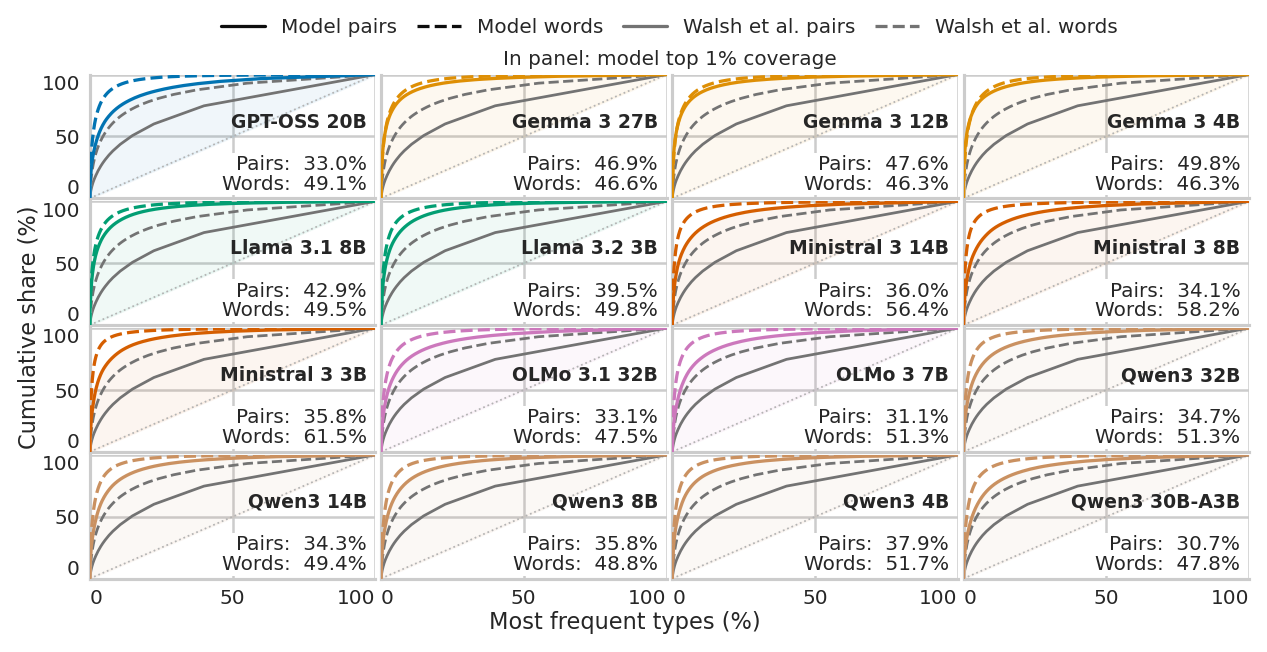}
    \caption{Cumulative concentration curves showing the cumulative share of document occurrences accounted for by the most frequent types. Lines closer to the top-left indicate greater collapse. Solid lines show rhyming pairs ($\leq 2$ line threshold), and dashed lines show content words. As an approximate human baseline, we include pooled results from the Poetry Foundation dataset by \citet{walshSonnetNotBot2024} (gray). In each cell, we also report the cumulative share of occurrences accounted for by the top 1\% of word types (\citeauthor{walshSonnetNotBot2024} Pairs: 11.87\%, Words: 24.89\%).}
    \label{fig:cumulative-concentration}
\end{figure*}

\subsection{Cumulative Concentration of Word and Rhyme Types}
Despite all models' demonstrated ability to employ a large number of valid rhymes, we find that, like non-rhyming vocabulary, the distributions of rhyming pairs are highly skewed towards a very small number of distinct types. Following prior work investigating poetic vocabulary richness in Slovak poetry \citep{Popescu2011VocabularyRI}, we show this disproportional word and rhyme utilization using per-model cumulative concentration curves in Figure \ref{fig:cumulative-concentration}. These curves show the cumulative percentage of document occurrences accounted for by the most frequent types. We also report the cumulative share of document occurrences covered by the top 1\% most frequent words and rhyming pairs for each model. As an approximate human-authored baseline, we visualize the same cumulative concentration curves for human poems in the Poetry Foundation corpus assembled by \citet{walshSonnetNotBot2024}. We emphasize here that the \citeauthor{walshSonnetNotBot2024} data is not balanced with respect to form or tag and is orders of magnitude smaller than our dataset (1,442 poems versus 1.1 million). We would also expect that the Poetry Foundation, being a collection of published poetry, lies at the far high end of human creative and poetic capability. However, they are still of interest as canonical references.

These cumulative concentration curves show how models rely heavily on a small fraction of their most frequent words and rhyming pairs. Across models, the top 1\% of pairs account for between 30.7\% (Qwen3 30B-A3B) and 49.8\% (Gemma 3 4B) of poem-pair occurrences. Content words are even more heavily concentrated among frequent types, with the top 1\% of types accounting for between 46.3\% (Gemma 3 4/12B) and 61.5\% (Ministral 3 3B) of poem occurrences. 

\begin{figure*}[ht]
    \centering
    \includegraphics[width=1\textwidth]{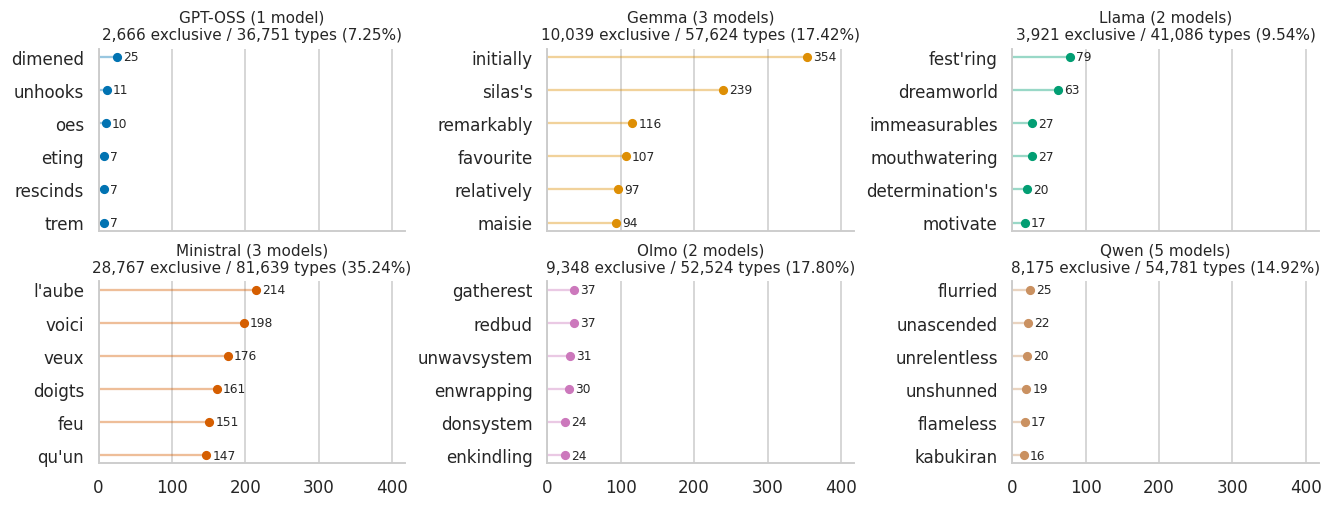}
    \caption{Model-unique vocabulary (word types appearing only in poems by a single model) grouped by family. We report the number and total vocabulary share of model-unique types, as well as the unique types appearing in the most poems per model family. Values on the $x$ axis indicate the total number of poems where a given type appears.} 
    \label{fig:family-unique}
\end{figure*}

\subsection{Model-Specific Vocabulary Signatures}\label{sec:model-specific}
Finally, a surprising feature of our dataset is the presence of non-trivial \textit{model-specific vocabularies}, or word types that  appear only in poems generated by a single model or model family. In Figure \ref{fig:family-unique}, we report the number and total vocabulary share of unique word types for each model family, as well as its most frequent unique word types. 

Certain types surfaced in this analysis necessitated further manual investigation. GPT-OSS, which has the smallest model-unique vocabulary (both absolutely and as a share of total vocabulary), would occasionally insert dashes or Markdown to make poetic structures more obvious: ``eting'', ``oes'', and ``trem'' are processing cruft corresponding to ``fle-eting'', ``ech-oes'', and ``trem-ble''. However, ``dimened'' is either a model misspelling of ``dimmed'' or a model neologism (``in the hush of a dimened room''). OLMo's ``unwavsystem'' and ``donsystem'' appear to be the result of sampling the ``system'' token in place of an apostrophe (``unwav'ring'', ``don't''). And, perhaps most obviously, we find that Ministral models rely heavily on French vocabulary, which accounts for its large share of model-unique words (35\%). 

Other high-frequency model-unique vocabulary offers some other surprising insights. For example, Gemma is the only family to ever use the commonplace adverbs ``initially'', ``remarkably'', or ``relatively''; it is also the only user of the British English spelling of ``favourite''. Qwen uses the malapropism ``unrelentless'' in 20 different poems, as well as the somewhat awkward ``unascended'' and ``unshunned'' almost as many times.\footnote{For a per-model breakdown of unique terms, see Figure \ref{fig:model-unique} in Appendix \ref{appendix:unique}.}

\section{Discussion}
\paragraph{All models collapse, but in different ways.} Our analyses point to both inter- and intra-model lexical homogenization. All models show high levels of lexical homogenization, but certain poetic forms exhibit greater collapse than others: for example, haiku are much more self-similar than couplets. Likewise, certain word types like ``light'' or ``world'' are very common in poems generated by any model, whereas other words are symptomatic of collapse for only select families (such as Gemma's ``plea'' or Ministral's ``oh''). While all models we studied exhibit creative collapse, the manner by which it manifests varies considerably both across and within model families. This suggests researchers can more precisely detect mode collapse with a detailed taxonomy to track against.

\paragraph{Language models are capable (but lazy) rhymers.} A surprising result of our analyses is the size of observed LLM rhyming vocabularies relative to their frequency. When reading LLM-generated rhyming poetry, it can seem as though the model has only memorized a few stock rhyming pairs; we even initially doubted whether models could actually rhyme at all. However, sampling poems at scale reveals that even very small models are able to rhyme thousands of distinct words, but that only 1\% of rhyming pairs account for between 30 and 50\% of observed rhymes. Moreover, we note that while the number of unique rhyming words and pairs is strongly correlated with model size, rhyming vocabulary as a share of total vocabulary is not. This suggests that rhyming ability is not necessarily constrained by parameter count at the 3B+ scale. Collapsed rhyming behavior should therefore be treated as a distinct problem from whether or not a model can rhyme a diverse set of words. 

\paragraph{Mode collapse obscures latent abilities.} While our large-scale analysis demonstrates that LLMs exhibit strong lexical characteristics of mode collapse, we also observe that these models possess a large inventory of rarer words and rhymes. At the far end of this ``long tail'', we note that all model families possess unique word types, meaning they appear in poems generated by one family alone. While some of these types reflect other failure modes (such as text formatting or non-English words), other families' model-unique vocabularies evidence more interesting behavior. We observe possible neologisms and malapropisms (``dimened'', ``unrelentless'') alongside surprisingly common words that are only used by a single family (``initially'', ``remarkably'', ``relatively''). This further supports our conclusion that while mode collapse is shared across LLMs, the precise lexical hallmarks vary considerably across models and families. Aggregate measures of mode collapse, particularly on small-scale corpora, can potentially elide these inter-model differences. More critically, they can obscure the latent capabilities LLMs possess in distinct domains, leading to poor ability estimations.

\section{Conclusion}
We demonstrate that, while LLMs often generate repetitive poems for common poetic forms and content topics, observing ca.\ 1.1 million poems from 16 contemporary open-weight models suggests this does not mean they do not learn \textit{how} to write a novel poem. We conclude that mode collapse can be better understood as \textit{model laziness}: models can produce diverse content, yet they avoid doing so. Sampling at scale exposes this latent ability often obscured by homogenization measures.

The literature commonly considers mode collapse to be a catastrophic failure type \citep{gerstgrasserModelCollapseInevitable2024,shumailovAIModelsCollapse2024}.
We disagree.
Our work bolsters a growing number of parallel efforts suggesting that even ``collapsed''  models are capable of emitting novel samples \citep{troshinControlTemperatureSelective2025,zhangVerbalizedSamplingHow2026} by demonstrating this behavior emerges even without directly manipulating the model or prompt to encourage it.

But recoverable novelty is difficult to demonstrate when the total range of possible outputs is unknown.
Our work thus closes a critical understanding gap by demonstrating that models are capable of emitting a diverse range of outputs, approaching human diversity for especially rare or unusual prompts.
This suggests those prompting models for diverse content should ensure they request unique content. That is, users should not leave their requests ambiguous or unconstrained when the goal is to yield non-memorized content.

\paragraph{Next steps.} Our work indicates models undergo mode collapse in more complex ways than previously understood.
One future avenue is to replicate and extend our method into other genres of text, e.g., storytelling or essay writing, which will require novel methods for describing the space of possibility in more unconstrained creative domains.

\section*{Limitations}
While our study entertains a large number of combinations over numerous poetic forms, topics, and models; our study remains limited to only a single language (English) and a particularly canonical understanding of what poetry is (as captured by the Poetry Foundation corpus). Patterns in poetic output may differ substantially in other languages or newer forms of poetry. Moreover, our generation procedure relied exclusively on naturalistic, zero-shot prompting with fixed temperature and sampling hyperparameters. This may materially impact lexical diversity in output. Setting temperature to 0 (for example) may reveal different characteristics. We furthermore make no claim about the formal adherence or aesthetic quality of generated poems, nor do we explore how language models exhibit mode collapse in non-poetic creative mediums.
Our claims hold exclusively to the specific models we tested, in English poetry only. 

\section*{Ethical Considerations}
Poetry is a deeply human form of art that has been practiced by cultures around the world for millennia.
The possibility of using LLMs and other AI systems to automatically produce creative content is a legitimate threat for all persons practicing art.
Consequently, any research investigating and comparing LLM capability gaps in creative domains versus humans implicitly (or explicitly) enables improvements to these systems, and thus any potential negative downstream impacts to livelihood, expression, and so on.
Our work involves generating  poetry because poetry exhibits greater structural regularity than prose, making it a useful base case for observing mode collapse.
We explicitly do not want to enable language models to be better at producing poetry, and so we do not release the ca. 1.1 million poems generated over the course of this project to prevent any potential misuse.

\section*{Acknowledgments}
We acknowledge the support of the Natural Sciences and Engineering Research Council of Canada (NSERC). Nous remercions le Conseil de recherches en sciences naturelles et en génie du Canada (CRSNG) de son soutien.


\bibliography{custom}

\appendix

\section{Dataset Metaprompt}
\label{sec:appendix-metaprompt}

We insert each form-tag pair into the following ``metaprompt'' template:

\begin{lstlisting}[
    basicstyle=\footnotesize\ttfamily,
    breaklines=true,
    breakatwhitespace=true,
    columns=fullflexible,
    keepspaces=true,
    showstringspaces=false,
    frame=single,
    framesep=5pt,
    linewidth=\columnwidth
]
Given the subject/theme/tag and the poetic form
provided below, write a clear prompt that asks
another LLM to compose an original poem. The 
prompt should explicitly request the specified
poetic form and that the poem focus on the given
subject/theme/tag. If the form is 'poem', then
you should just request a poem without any
particular formal specifications. Finally, your
prompt should ask the model to generate only the
poem, without any additional text or 
information.

**Inputs:**
    Subject/theme/tag: {tag}
    Poetic form: {form}
**Output format**:
    Provide only the generated poem-request prompt.
    Do not generate any other text besides the prompt.
\end{lstlisting}

\section{Preprocessing}\label{appendix:preprocessing}
Despite the explicit instructions in our metaprompt, models frequently produced titles, summaries, rhyme scheme annotations, markdown, and other extraneous text in addition to generating poetry. Following Unicode normalization, we use regular expressions to identify and remove obvious instances of the following types of non-poetic text: 
\begin{itemize}
    \item introductory commentary on the prompt,
    \item refusals,
    \item titles,
    \item trailing commentary after the poem is completed,
    \item formatting such as Markdown or HTML tags, and
    \item decorations or annotations, such as rhyme scheme or Roman numerals.
\end{itemize}

\begin{table}[h]

    \centering
    \begin{tabular}{lc}
        Operation & \% \\
        \hline
        Any cleaning & 38.65 \\
        Title & 28.55 \\
        Introduction & 3.41 \\
        Postscript & 4.82 \\
        Decoration/annotation & 23.49 \\
        Refusal & 0.03 \\
        \bottomrule
    \end{tabular}
    \caption{Percent of generated poems affected by data cleaning operations.}
    \label{tab:cleaning}
\end{table}
We report the impact of our cleaning procedures in Table \ref{tab:cleaning}. The most commonly performed cleaning operation was title removal, which we analyze in the following section.

\section{Titles}\label{appendix:titles}
\begin{figure}[ht]
    \centering
    \includegraphics[width=1\linewidth]{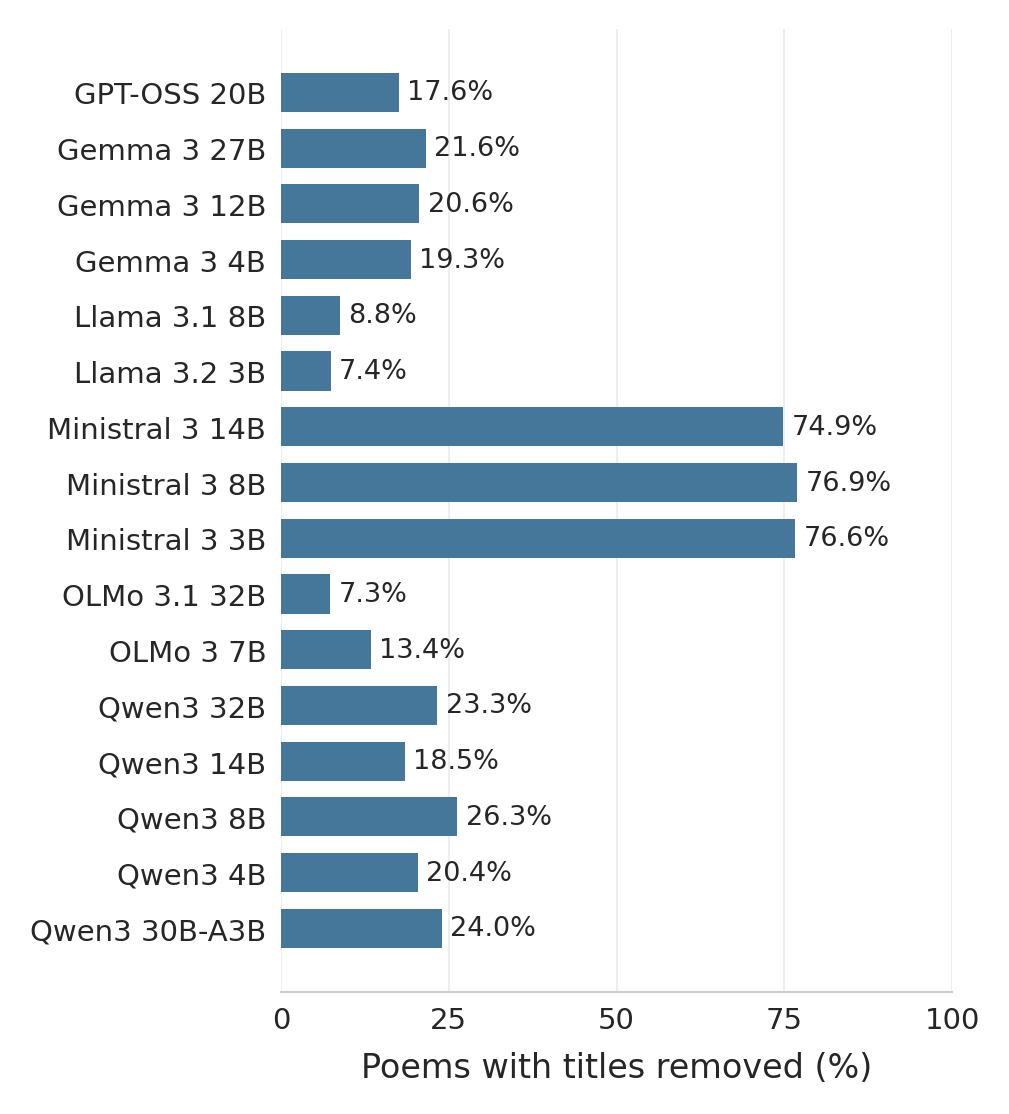}
    \caption{Percent of poems with titles removed during cleaning per model.}
    \label{fig:title-stats}
\end{figure}

While titles were not requested in our metaprompt, we find that all models occasionally generate poem titles, and some models even generate titles for most poems. Though we do not treat titles in our primary analyses, titles still constitute generated text. As such, we report the number of poems with detected titles per generator model in Figure \ref{fig:title-stats}.

We detect titles in at least 74.9\% of poems generated by Ministral models. This is much higher than any other model or family: Qwen3 8B ranks closest with only 26.3\% of titles.

\begin{table*}[h]
    \small
    \centering
    \begin{tabular*}{\textwidth}{@{\extracolsep{\fill}}lccr@{}}
        Title & Poems & Models & Top Generator (count) \\
        \hline
        The Weight of Light & 1,219 & 11 & Ministral 3 3B (426) \\
        The Weight of What Was & 1,042 & 10 & Ministral 3 14B (441) \\
        The Weaver's Thread & 953 & 14 & Ministral 3 14B (265) \\ 
        Elegy for the Unseen & 837 & 13 & Qwen3 30B-A3B (255) \\
        The Weight of Wings & 803 & 14 & Ministral 3 8B (321) \\
        The Weight of Hands & 714 & 10 & Ministral 3 3B (273) \\
        The Weight of Silence & 539 & 15 & Ministral 3 3B (141) \\ 
        The Weight of Echoes & 490 & 11 & Ministral 3 3B (246) \\
        Static Bloom & 482 & 3 & Gemma 3 27B (337) \\ 
        The Weight of the Unseen & 443 & 8 & Ministral 3 14B (183) \\
        \bottomrule
    \end{tabular*}
    \caption{Top-10 most-repeated titles generated by any model.}
    \label{tab:top-titles}
\end{table*}

Despite the large number of generated titles across models, \textit{the majority of titles appear more than once}. In other words, even poem titles exhibit mode collapse, though this collapse differs by model. We find that 32,485 titles map to 195,153 poems, and that 20,684 titles are generated by multiple models. In Table \ref{tab:top-titles}, we present the top-10 most-repeated titles generated by any model. These titles evince some of the same lexical patterns identified in Table \ref{fig:most-common-output-words} (``light'', ``silence''). We also note that most of the top Ministral-dominated titles are just variations on ``The Weight of ...''. 

\begin{figure}
    \centering
    \includegraphics[width=1\linewidth]{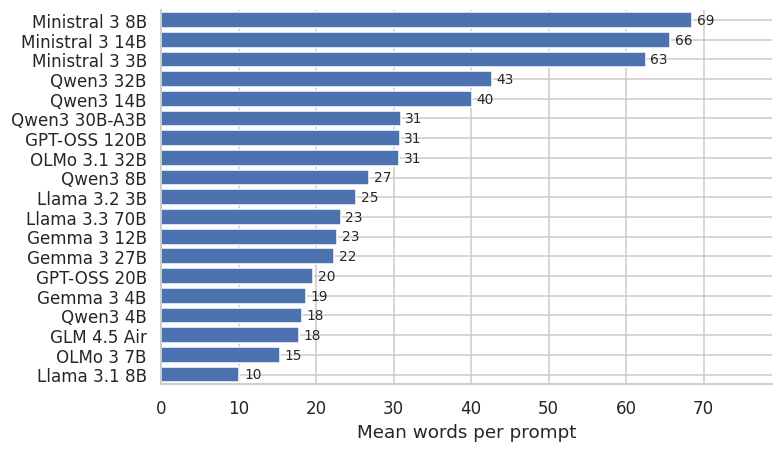}
    \caption{Mean word length of generated prompts by model.}
    \label{fig:prompt-length}
\end{figure}

\section{Prompt Length}
While we do not report a comprehensive analysis of our generated prompts, we note substantial variance in prompt generation behavior across models, particularly with respect to verbosity. We provide the mean length of generated prompts in Figure \ref{fig:prompt-length}. Ministral models are most verbose, averaging between 63 and 69 words per prompt; Llama 3.1 8B is most concise, averaging only 10.

\section{Rhyme Vocabulary}\label{appendix:rhyme-vocab}
\begin{figure*}
    \centering
    \includegraphics[width=\textwidth]{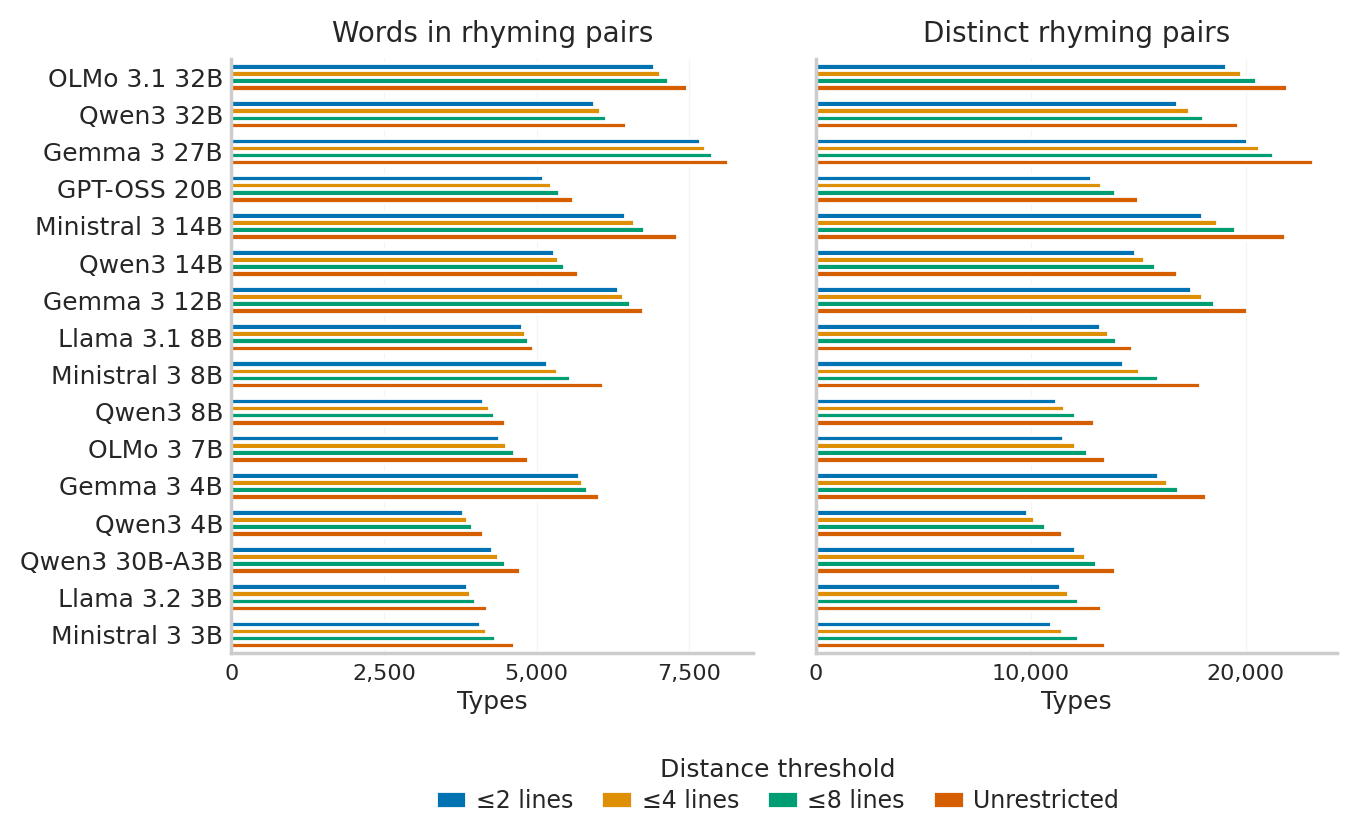}
    \caption{Distinct words present in line-terminal rhyming pairs (left), and total distinct rhyming pairs (right). Color corresponds to the maximum line distance rule for counting pairs.}
    \label{fig:rhyme-vocab}
\end{figure*}
In Figure \ref{fig:rhyme-vocab}, we show the number of unique content words present in rhyming pairs as well as the total number of unique rhyming pairs. Models are listed by size in descending order, showing the correlation between model size and both measures of rhyming capacity. 

\section{Unique Vocabulary}\label{appendix:unique}
\begin{figure*}
    \centering
    \includegraphics[width=\textwidth]{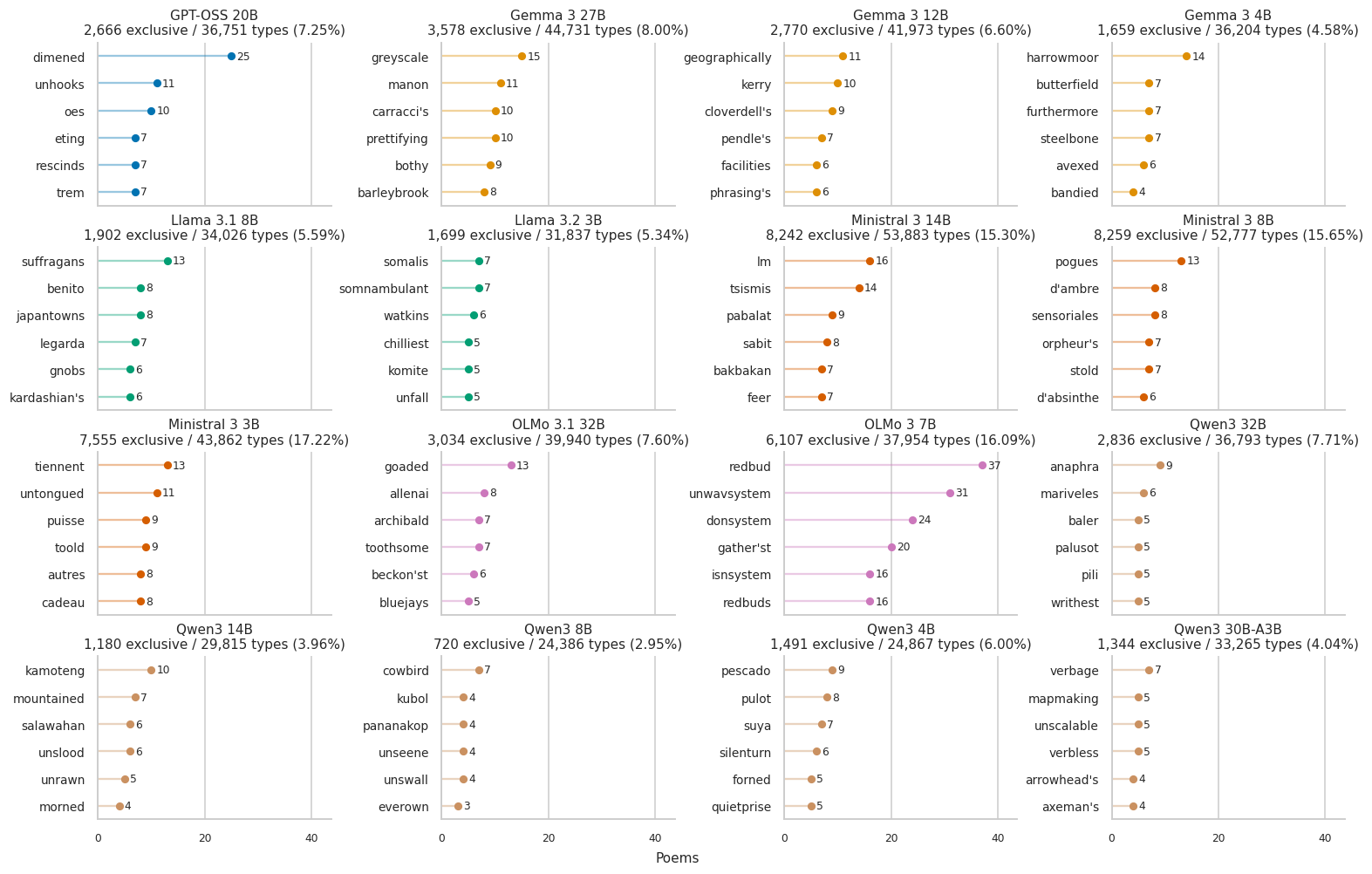}
    \caption{Most frequent words appearing uniquely in poems generated by a given model. Beneath each model name, we also report the total number of model-unique words observed for that model.}
    \label{fig:model-unique}
\end{figure*}
To complement our family-level analysis in Section \ref{sec:model-specific}, we present the most-frequent model-unique vocabulary per generator model in Figure \ref{fig:model-unique}. 

\end{document}